\documentclass[10pt,twocolumn,letterpaper]{article}

\usepackage{cvpr}
\usepackage{times}
\usepackage{epsfig}
\usepackage{graphicx}
\usepackage{amsmath}
\usepackage{amssymb}

\usepackage{multirow}
\usepackage{mathrsfs}
\usepackage{enumitem}
\usepackage{makecell}
\usepackage{tabulary}
\usepackage{pifont}
\usepackage{subfigure}
\usepackage{bm}

\usepackage[breaklinks=true,bookmarks=false]{hyperref}

\cvprfinalcopy % *** Uncomment this line for the final submission

\def\cvprPaperID{9683} % *** Enter the CVPR Paper ID here

\ifcvprfinal\fi
\begin{document}

%%%%%%%%% TITLE
\title{X-Linear Attention Networks for Image Captioning}

\author{Yingwei Pan, Ting Yao, Yehao Li, and Tao Mei \\
{\normalsize\centering JD AI Research, Beijing, China}\\
{\tt\small \{panyw.ustc, tingyao.ustc, yehaoli.sysu\}@gmail.com, tmei@jd.com}
}

\maketitle
\thispagestyle{empty}

%%%%%%%%% ABSTRACT

\begin{abstract}
Recent progress on fine-grained visual recognition and visual question answering has featured Bilinear Pooling, which effectively models the 2$^{nd}$ order interactions across multi-modal inputs. Nevertheless, there has not been evidence in support of building such interactions concurrently with attention mechanism for image captioning. In this paper, we introduce a unified attention block --- X-Linear attention block, that fully employs bilinear pooling to selectively capitalize on visual information or perform multi-modal reasoning. Technically, X-Linear attention block simultaneously exploits both the spatial and channel-wise bilinear attention distributions to capture the 2$^{nd}$ order interactions between the input single-modal or multi-modal features. Higher and even infinity order feature interactions are readily modeled through stacking multiple X-Linear attention blocks and equipping the block with Exponential Linear Unit (ELU) in a parameter-free fashion, respectively. Furthermore, we present X-Linear Attention Networks (dubbed as X-LAN) that novelly integrates X-Linear attention block(s) into image encoder and sentence decoder of image captioning model to leverage higher order intra- and inter-modal interactions. The experiments on COCO benchmark demonstrate that our X-LAN obtains to-date the best published CIDEr performance of 132.0\% on COCO Karpathy test split. When further endowing Transformer with X-Linear attention blocks, CIDEr is boosted up to 132.8\%. Source code is available at \url{https://github.com/Panda-Peter/image-captioning}.
\end{abstract}

\begin{figure}[!tb]
\vspace{-0.28in}
\centering {\includegraphics[width=0.46\textwidth]{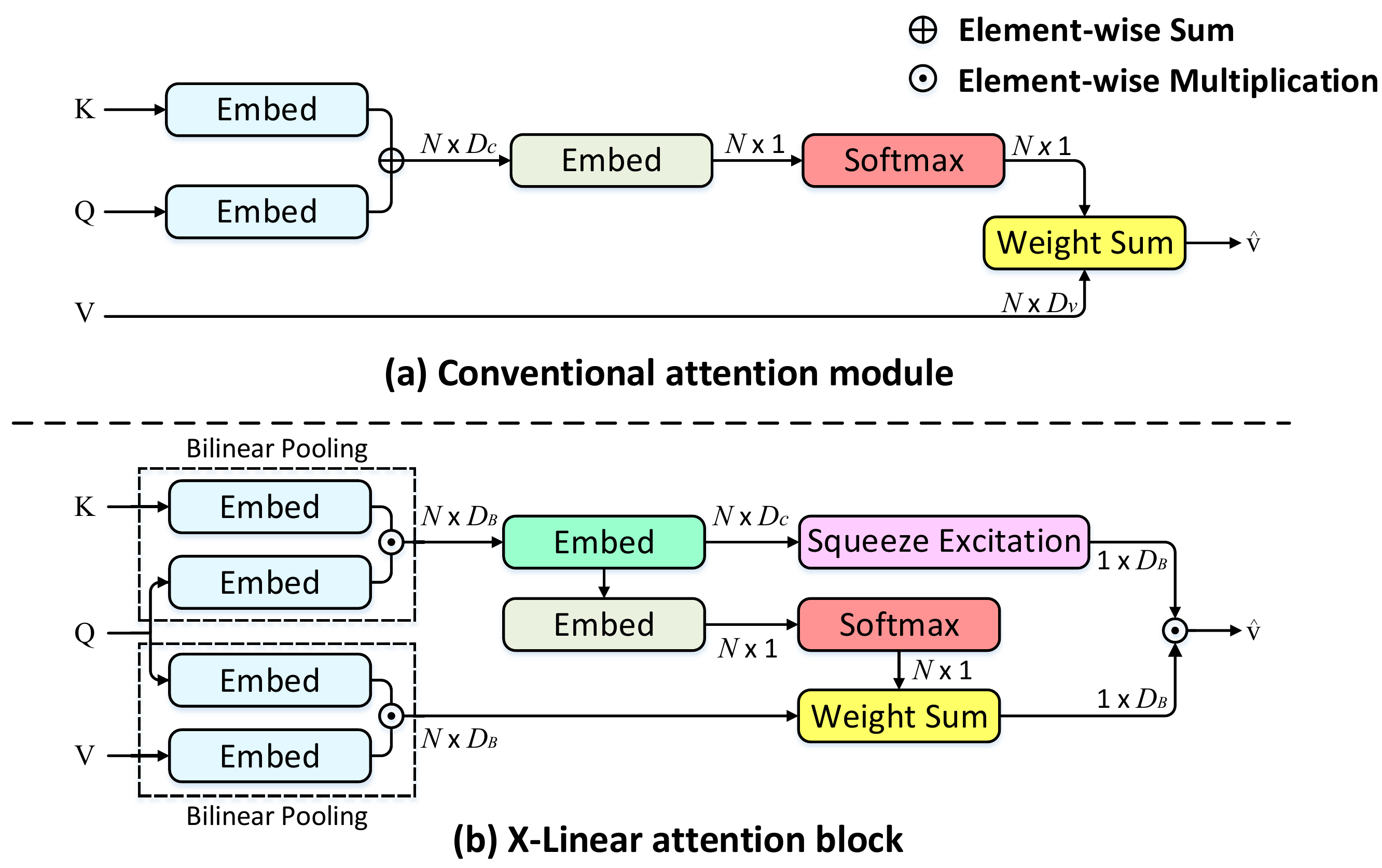}}
\vspace{-0.06in}
\caption{Comparison between conventional attention mechanism and our X-Linear attention block for image captioning. (a) Conventional attention mechanism linearly fuses query (Q) and key (K) via element-wise sum to compute spatial attention weight for each value (V), which characterizes the 1$^{st}$ order interaction between query and key. (b) X-Linear attention block fully capitalizes on bilinear pooling to capture the 2$^{nd}$ order feature interaction in between, and measures both spatial and channel-wise attention distributions. The two attention weights are adopted to accumulate the enhanced values of bilinear pooling on query and value.}
\label{fig:fig1}
\vspace{-0.2in}
\end{figure}

\vspace{-0.2in}
\section{Introduction}
Image captioning is the task of automatically producing a natural-language sentence to describe the visual content of an image. The essential practice of neural captioning models follows encoder-decoder paradigm \cite{Mao:NIPS14,Vinyals14}, which is derived from neural machine translation \cite{Sutskever:NIPS14}. In between, Convolutional Neural Network (CNN) is utilized to encode an input image and Recurrent Neural Network (RNN) is adopted as sentence decoder to generate the output sentence, one word at each time step. Despite involving two different major modalities (visual content and textual sentence) in image captioning, such paradigm of approaches seldom explores the multi-modal interactions particularly at the early stage. In other words, vision and language are treated independently. That prompts the recent state-of-the-art methods \cite{anderson2017bottom,Xu:ICML15} to adopt visual attention mechanisms which trigger the interaction between visual content and natural sentence. Concretely, these visual attention mechanisms boost performance by learning to identify selective spatial regions in an image conditioning on current hidden state of language decoder, and in turn accumulating encoded region features with attention weights to guide decoding process. Figure \ref{fig:fig1}(a) illustrates the most conventional attention measure which estimates attention weights via linearly fusing the given query (hidden state of sentence decoder) and key (encoded image features) from different modalities. The attention is then applied to the value (encoded image features) to derive a weighted sum. Nevertheless, we argue that the design of conventional attention inherently exploits only the 1$^{st}$ order feature interaction and is still lacking in efficacy. That severely limits the capacity of complex multi-modal reasoning in image captioning.

A natural way to mitigate the problem is to capture higher order interactions. We start our exploration from 2$^{nd}$ order interaction through the unique design of a unified attention block, namely X-Linear attention block, as shown in Figure \ref{fig:fig1}(b). Technically, the outer product of key and query is computed through bilinear pooling to take all pairwise interactions between query and key into account. After bilinear pooling, two embedding layers are exploited to predict attention weights for each spatial region, followed by a softmax layer to normalize the spatial attention vector. In the meantime, the embedded outer product (feature map) is passed through a squeeze-excitation operation. The squeeze operation aggregates the feature map across spatial regions to produce a channel descriptor and the excitation operation performs the self-gating mechanism with a sigmoid on the channel descriptor to obtain the channel-wise attention vector. Finally, the outer product of query and value via bilinear pooling is weighted summated with the spatial attention vector, and we take the channel-wise multiplication of the sum and the channel-wise attention vector as the attended features. As such, our X-Linear attention block builds the 2$^{nd}$ order interactions and infers the joint representations for image features and hidden states. It is also appealing in view that a stack of the blocks is readily grouped to go beyond bilinear models and extract higher order interactions. In the extreme case, our model could create infinity order interactions by stacking numerous X-Linear attention blocks and we implement this via the kernel trick, e.g., Exponential Linear Unit (ELU), in practice.

By integrating X-Linear attention block(s) into image captioning structures, we present a new X-Linear Attention Networks (X-LAN) to leverage high order intra- and inter-modal interactions, respectively, in the encoder and decoder. Specifically, for image encoder, Faster R-CNN is firstly utilized to detect a set of image regions. After that, a stack of X-Linear attention blocks are adopted to encode the region-level features with the higher order intra-modal interaction in between, leading to a set of enhanced region-level and image-level features. Conditioned on the enhanced visual features induced by image encoder, we further employ X-Linear attention block in sentence decoder to perform multi-modal reasoning. This encourages the exploration of high order inter-modal interactions between visual content and natural sentence to boost sentence generation.

The main contribution of this work is the proposal of a unified X-Linear attention block that models the 2$^{nd}$ order interactions with both spatial and channel-wise bilinear attention. This also leads to the elegant view of how the block should be extended for mining higher or even infinity order interactions and how to integrate such block(s) into image captioning structure. Through an extensive set of experiments, we demonstrate that our new X-LAN model achieves new state-of-the-art performances on COCO dataset.

\section{Related Work}\label{sec:RW}

\textbf{Image Captioning.} Image captioning is an active research area \cite{anderson2017bottom,huang2019attentio,li2019novel,Xiong2016MetaMind,Mao:NIPS14,rennie2017self,Vinyals14,wang2019paragraph,Xu:ICML15,yao2017novel,yao2019hierarchy,yao2017boosting,You:CVPR16}. The early attempts \cite{Mao:NIPS14,Vinyals14} exploit the encoder-decoder paradigm that firstly utilizes CNN to encoder image and then adopts RNN based decoder to generate the output word sequence, leading to promising results for this task. After that, a series of innovations have been proposed to boost image captioning by encouraging more interactions between the two different modalities via attention mechanism \cite{cho2015describing}. In particular, \cite{Xu:ICML15} integrates soft and hard attention mechanism into LSTM based decoder, aiming to select the most relevant image regions for word prediction at each decoding stage. \cite{You:CVPR16} presents semantic attention that learns to selectively focus on the semantic attributes in image for sentence generation. Instead of fully performing visual attention as in \cite{Xu:ICML15}, \cite{Xiong2016MetaMind} proposes an adaptive attention model that dynamically decides whether to attend to image regions at each decoding stage. Furthermore, bottom-up and top-down attention mechanism \cite{anderson2017bottom} exploits visual attention at object level via bottom-up mechanism, and all salient image regions are associated with the output words through top-down mechanism for image captioning. \cite{qin2019look} presents the look back method to integrate attention weights from previous time step into the measurement of attention at current time step, which suits visual coherence of human. Later on, the most recently proposed attention on attention module \cite{huang2019attentio} enhances visual attention by further measuring the relevance between the attention result and the query.

Much of existing attention mechanisms in image captioning have concentrated on the exploration of only the 1$^{st}$ order feature interaction between image content and sentence, reflecting limited capacity of multi-modal reasoning. In contrast, we design a novel X-Linear attention block to capture higher and even infinity order interactions, which facilitate both single-modal feature enhancement and multi-modal reasoning for image captioning.

\textbf{Bilinear Pooling.} Bilinear pooling is an operation to calculate outer product between two feature vectors. Such technique can enable the 2$^{nd}$ order interaction across all elements in feature vectors and thus provide more discriminative representations than linear pooling. An early pioneering work \cite{lin2015bilinear} demonstrates the advantage of bilinear pooling for fine-grained visual recognition task. Local pairwise feature interactions are thus modeled by leveraging bilinear pooling over the outputs of two CNNs. Later on, \cite{gao2016compact} proposes compact bilinear pooling that efficiently compresses the high-dimensional bilinear pooling feature into compact one with a few thousand dimensions, but retains the same discriminative power in the meantime. \cite{fukui2016multimodal} further extends compact bilinear pooling into multi-modal scenario where visual and textual representations are combined for visual question answering task. Instead of compact bilinear pooling that needs complex computations, \cite{kim2016hadamard} proposes a flexible low-rank bilinear pooling structure with linear mapping and Hadamard product. Recently, \cite{yu2018hierarchical} presents a hierarchical bilinear pooling model to aggregate multiple cross-layer bilinear pooling features for fine-grained visual recognition. \cite{kim2018bilinear} exploits low-rank bilinear pooling to construct bilinear attention network, aiming to learn bilinear attention distributions for visual question answering.

The aforementioned bilinear pooling techniques are mainly designed for fine-grained visual recognition or visual question answering. Instead, our X-Linear attention block is applicable to image encoder and sentence decoder to exploit higher order intra and inter-modal interactions for image captioning task.

\section{X-linear Attention Networks (X-LAN)}

In this section, we introduce a novel unified formulation of attention module, named X-Linear attention block, that fully capitalizes on bilinear pooling to capture the 2$^{nd}$ order feature interactions with both spatial and channel-wise bilinear attention. Moreover, we show a specific integration of X-Linear attention block into image encoder and sentence decoder to capture higher order intra- and inter-modal interactions, aiming to enhance visual information and perform complex multi-modal reasoning for image captioning.

\subsection{Conventional Attention Module}
We first provide a brief review of the most conventional attention module \cite{Xu:ICML15} applied in image captioning, which learns to selectively attend to salient image regions for sentence generation. Formally, at decoding time step $t$, conditioned on the query ${\bf{Q}}$ (current hidden state of sentence decoder ${\bf{h}}_t$), we can obtain the attention distribution ${\bm{\alpha}}^t$ over a set of keys ${\bf{K}}=\{{\bf{k}}_i\}_{i=1}^N$ ($N$ local image features):
\begin{equation}\label{Eq:Eq1}\small
a^t_{i}={\bf{W}}_a\left[\tanh\left({\bf{W}}_k{{\bf{k}}_i} + {\bf{W}}_q{\bf{Q}}\right)\right],{\bm{\alpha}}^t=softmax\left({\bf{a}}^t\right),
\end{equation}
where ${\bf{W}}_a$, ${\bf{W}}_k$, and ${\bf{W}}_q$ are embedding matrices, and $a^t_{i}$ denotes the $i$-th element in ${\bf{a}}^t$.
In this sense, the normalized attention weight $\alpha^t_{i}$ for each local image feature ($i$-th element in ${\bm{\alpha}}^t$) is derived from the linear fusion of the given query and key via element-wise sum. Such way inherently exploits only the 1$^{st}$ order feature interaction between natural sentence and visual content for attention measurement. Next, attention module produces the attended image feature $\hat{\bf{v}}^t$ by accumulating all values ${\bf{V}}=\{{\bf{v}}_i\}_{i=1}^N$ ($N$ local image features) with spatial attention weights: $\hat{\bf{v}}^t  = \sum\nolimits_{i = 1}^N {\alpha^t_{i}{\bf{v}}_i}$.

\subsection{X-Linear Attention Block}
Though conventional attention module nicely triggers the interaction between different modalities, only the 1$^{st}$ order feature interaction is exploited, which reflects limited capacity of complex multi-modal reasoning in image captioning.
Inspired by the recent successes of bilinear pooling applied in fine-grained visual recognition \cite{gao2016compact,yu2018hierarchical} or visual question answering \cite{fukui2016multimodal,kim2018bilinear}, we fully capitalize on bilinear pooling techniques to construct a unified attention module (X-Linear attention block) for image captioning, as depicted in Figure \ref{fig:fig1}(b). Such design of X-Linear attention block strengthens the representative capacity of the output attended feature by exploiting higher order interactions between the input single-modal or multi-modal features.

In particular, suppose we have the query ${\bf{Q}}\in {\mathbb{R}}^{D_q}$, a set of keys ${\bf{K}}=\{{\bf{k}}_i\}_{i=1}^N$, and a set of values ${\bf{V}}=\{{\bf{v}}_i\}_{i=1}^N$, where ${\bf{k}}_i \in {\mathbb{R}}^{D_k}$ and ${\bf{v}}_i \in {\mathbb{R}}^{D_v}$ denote the $i$-th key/value pair. X-Linear attention block firstly performs low-rank bilinear pooling to achieve a joint bilinear query-key representation ${\bf{B}}^k_i \in {\mathbb{R}}^{D_B}$ between query ${\bf{Q}}$ and each key ${\bf{k}}_i$:
\begin{equation}\label{Eq:Eq2}\small
{\bf{B}}^k_i=\sigma\left({\bf{W}}_k{{\bf{k}}_i}\right) \odot \sigma\left({\bf{W}}^k_q{\bf{Q}}\right),
\end{equation}
where ${\bf{W}}_k\in {\mathbb{R}}^{{D_B}\times{D_k}}$, and ${\bf{W}}^k_q\in {\mathbb{R}}^{{D_B}\times{D_q}}$ are embedding matrices, $\sigma$ denotes ReLU unit, and $\odot$ represents element-wise multiplication. As such, the learnt bilinear query-key representation ${\bf{B}}^k_i$ conveys the 2$^{nd}$ order feature interactions between query and key.

Next, depending on all bilinear query-key representations $\{{\bf{B}}^k_i\}_{i=1}^N$, two kinds of bilinear attention distributions are obtained to aggregate both spatial and channel-wise information within all values. Most specifically, the spatial bilinear attention distribution is introduced by projecting each bilinear query-key representation into the corresponding attention weight via two embedding layers, followed with a softmax layer for normalization:
\begin{equation}\label{Eq:Eq3}\small
{\bf{B}}^{'k}_i = \sigma\left({\bf{W}}^k_B{\bf{B}}^k_i\right), b^s_{i}={\bf{W}}_b{\bf{B}}^{'k}_i, {\bm{\beta}}^s=softmax\left({\bf{b}}^s\right),
\end{equation}
where ${\bf{W}}^k_B \in {\mathbb{R}}^{{D_c}\times{D_B}}$ and ${\bf{W}}_b \in {\mathbb{R}}^{1\times{D_c}}$ are embedding matrices, ${\bf{B}}^{'k}_i$ is the transformed bilinear query-key representation, and $b^s_{i}$ is the $i$-th element in ${\bf{b}}^s$. Here each element ${\bm{\beta}}_i^s$ in ${\bm{\beta}}^s$ denotes the normalized spatial attention weight for each key/value~pair. Meanwhile, we perform a squeeze-excitation operation \cite{hu2018squeeze} over all transformed bilinear query-key representations $\{{\bf{B}}^{'k}_i\}_{i=1}^N$ for channel-wise attention measurement. Concretely, the operation of squeeze aggregates all transformed bilinear query-key representations via average pooling, leading to a global channel descriptor $\bar{\bf{B}}$:
\begin{equation}\label{Eq:Eq4}\small
\bar{\bf{B}}={1 \over N}\sum\nolimits_{i = 1}^N {{\bf{B}}^{'k}_i}.
\end{equation}
After that, the followed excitation operation produces channel-wise attention distribution ${\bm{\beta}}^c$ by leveraging the self-gating mechanism with a sigmoid over the global channel descriptor $\bar{\bf{B}}$:
\begin{equation}\label{Eq:Eq5}\small
{\bf{b}}^c={\bf{W}}_e \bar{\bf{B}},{\bm{\beta}}^c=sigmoid\left({\bf{b}}^c\right),
\end{equation}
where ${\bf{W}}_e \in {\mathbb{R}}^{{D_B}\times{D_c}}$ is embedding matrix.

Finally, our X-Linear attention block generates the attended value feature $\hat{\bf{v}}$ by accumulating the enhanced bilinear values with spatial and channel-wise bilinear attention:
\begin{equation}\label{Eq:Eq6}\small
\begin{array}{l}
\hat{\bf{v}}=F_{X-Linear}\left({\bf{K}}, {\bf{V}}, {\bf{Q}}\right)={\bm{\beta}}^c\odot\sum\nolimits_{i = 1}^N {{\bm{\beta}}_i^s{{\bf{B}}^v_i}},\\
{\bf{B}}^v_i=\sigma\left({\bf{W}}_v{{\bf{v}}_i}\right) \odot \sigma\left({\bf{W}}^v_q{\bf{Q}}\right),
\end{array}
\end{equation}
where ${\bf{B}}^v_i$ denotes the enhanced value of bilinear pooling on query ${\bf{Q}}$ and each value ${{\bf{v}}_i}$, ${\bf{W}}_v\in {\mathbb{R}}^{{D_B}\times{D_v}}$, and ${\bf{W}}^v_q\in {\mathbb{R}}^{{D_B}\times{D_q}}$ are embedding matrices. Accordingly, compared to conventional attention modules that simply explore 1$^{st}$ order interaction between query and key, X-Linear attention block produces the more representative attended feature since higher order feature interactions are exploited via bilinear pooling.

\begin{figure}[!tb]
    \centering {\includegraphics[width=0.43\textwidth]{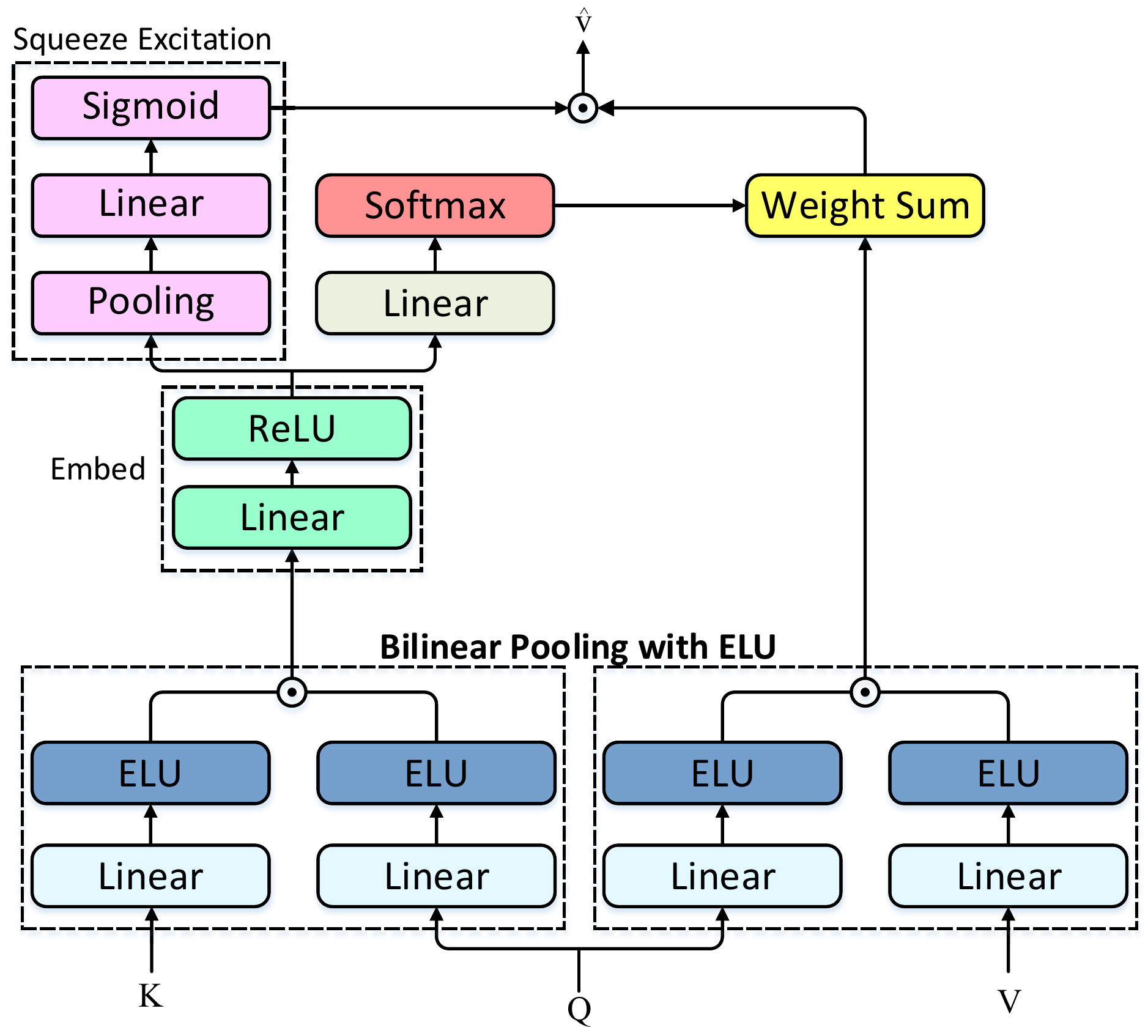}}
    \vspace{-0.1in}
    \caption{ A schematic diagram of X-Linear attention block plus ELU to capture infinity order feature interactions.}
    \label{fig:architecture}
    \vspace{-0.2in}
\end{figure}

\begin{figure*}[!tb]
\vspace{-0.2in}
    \centering {\includegraphics[width=0.93\textwidth]{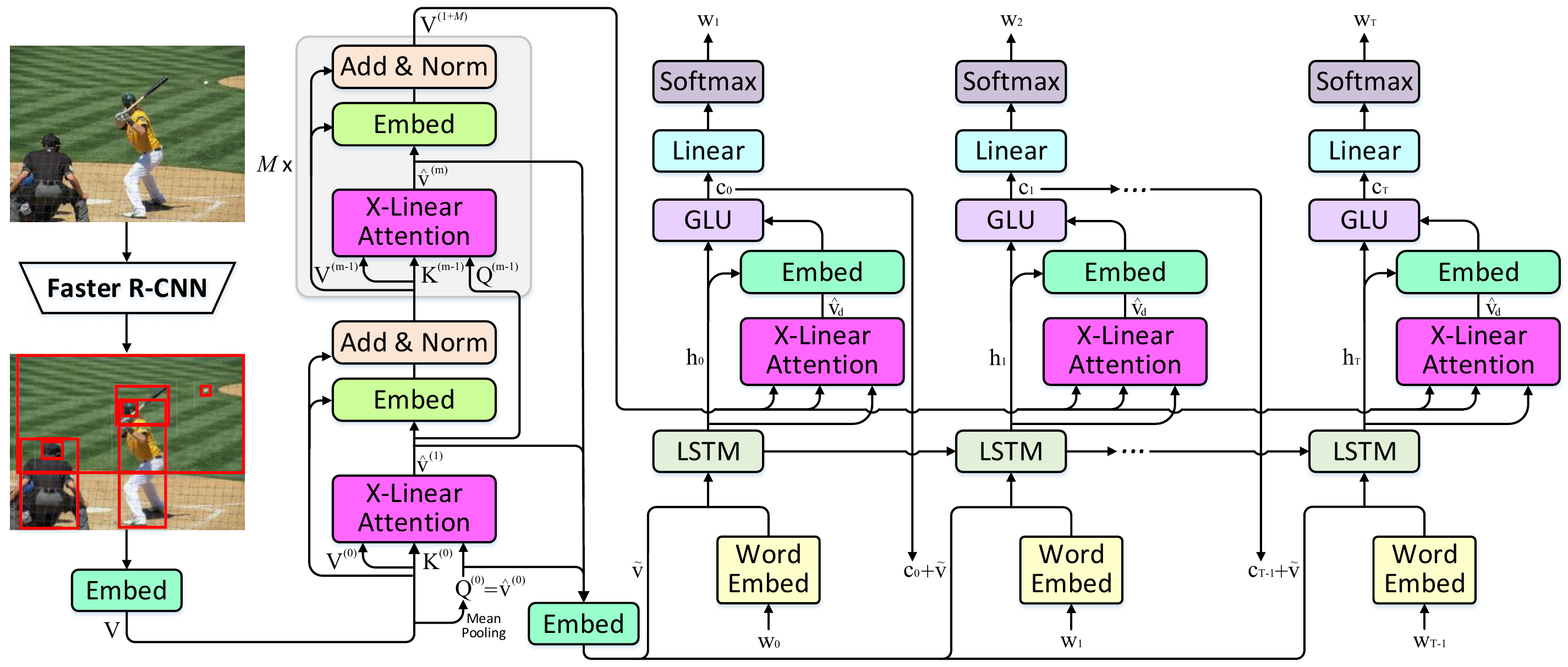}}
    \vspace{-0.06in}
    \caption{Overview of our X-Linear Attention Networks (X-LAN) for image captioning. Faster R-CNN is firstly utilized to detect a set of image regions. Next, a stack of X-Linear attention blocks are leveraged in image encoder to encode the region-level features with the higher order intra-modal interaction in between, leading to a set of enhanced region-level and image-level features. Depending on the enhanced visual features, X-Linear attention block is further adopted in sentence decoder to perform multi-modal reasoning. This encourages the exploration of high order inter-modal interactions between visual content and natural sentence to boost sentence generation.}
    \label{fig:framework}
    \vspace{-0.2in}
\end{figure*}

\textbf{Extension with higher order interactions.}
In order to exploit higher order feature interactions, we further iterate the above process of bilinear attention measurement and feature aggregation using a stack of our X-Linear attention blocks. Formally, for the $m$-th X-Linear attention block, we firstly take the pervious output attended feature $\hat{\bf{v}}^{(m-1)}$ as input query, coupled with current input keys ${\bf{K}}^{(m-1)}=\{{\bf{k}}_i^{(m-1)}\}_{i=1}^N$, and values ${\bf{V}}^{(m-1)}=\{{\bf{v}}_i^{(m-1)}\}_{i=1}^N$:
\begin{equation}\label{Eq:Eq7-1}\small
\hat{\bf{v}}^{(m)}=F_{X-Linear}\left( {\bf{K}}^{(m-1)}, {\bf{V}}^{(m-1)}, \hat{\bf{v}}^{(m-1)}\right),
\end{equation}
where $\hat{\bf{v}}^{(m)}$ is the output new attended feature. $\hat{\bf{v}}^{(0)}$, ${\bf{K}}^{(0)}$, and ${\bf{V}}^{(0)}$ denotes ${\bf{Q}}$, ${\bf{K}}$, and ${\bf{V}}$, respectively. Then, all keys/values are further updated conditioned on the output new attended feature $\hat{\bf{v}}^{(m)}$:
\begin{equation}\label{Eq:Eq7-2}\small
\begin{array}{l}
{\bf{k}}_i^{(m)}=LayerNorm(\sigma({\bf{W}}^k_m[\hat{\bf{v}}^{(m)},{\bf{k}}_i^{(m-1)}])+{\bf{k}}_i^{(m-1)}),\\
{\bf{v}}_i^{(m)}=LayerNorm(\sigma({\bf{W}}^v_m[\hat{\bf{v}}^{(m)},{\bf{v}}_i^{(m-1)}])+{\bf{v}}_i^{(m-1)}),
\end{array}
\end{equation}
where ${\bf{W}}^k_m$ and ${\bf{W}}^v_m$ are embedding matrices. Note that here each key/value is concatenated with the new attended feature, followed with a residual connection and layer normalization as in \cite{vaswani2017attention}. We repeat the process (Eq.(\ref{Eq:Eq7-1}) and Eq.(\ref{Eq:Eq7-2})) $M$ times via stacking $M$ X-Linear attention blocks, which captures higher ($2M^{th}$) order feature interactions.

\textbf{Extension with infinity order interactions.}
One natural way to exploit more higher (even infinity) order feature interactions is to stack plenty of X-Linear attention blocks. Nevertheless, such way inevitably leads to a huge rise in memory demand and computational cost, not to mention the extreme case of stacking infinity blocks. Instead, we adopt a simple but effective method to enable our X-Linear attention block to model infinity order interactions by additionally encoding query ${\bf{Q}}$, each key ${\bf{k}}_i$, and each value ${\bf{v}}_i$ with Exponential Linear Unit (ELU) \cite{barron2017continuously}, as shown in Figure \ref{fig:architecture}. That is, the infinity order feature interactions can be approximately modeled via performing bilinear pooling on two exponentially transformed features. Here we demonstrate that such approximation can be proved via Taylor expansion of each element in bilinear vector after exponential transformation. Specifically, given two feature vectors $X$ and $Y$, the Taylor expansion of bilinear pooling over the exponentially transformed features can be expressed as:
\begin{equation}\label{Eq:Eq8}\scriptsize
\begin{array}{l}
\exp ({{\rm{W}}_X}{\rm{X}}) \odot \exp ({{\rm{W}}_Y}{\rm{Y}}) \\
= [\exp ({\rm{W}}_X^1{\rm{X}}) \odot \exp ({\rm{W}}_Y^1{\rm{Y}}),...,\exp ({\rm{W}}_X^D{\rm{X}}) \odot \exp ({\rm{W}}_Y^D{\rm{Y}})]  \\
= [\exp ({\rm{W}}_X^1{\rm{X}} + {\rm{W}}_Y^1{\rm{Y}}),...,\exp ({\rm{W}}_X^D{\rm{X}} + {\rm{W}}_Y^D{\rm{Y}})]  \\
= [\sum\limits_{p = 0}^\infty  {\gamma _p^1{{({\rm{W}}_X^1{\rm{X}} + {\rm{W}}_Y^1{\rm{Y}})}^p}} ,...,\sum\limits_{p = 0}^\infty  {\gamma _p^D{{({\rm{W}}_X^D{\rm{X}} + {\rm{W}}_Y^D{\rm{Y}})}^p}} ],
\end{array}
\end{equation}
where ${\rm{W}}_X$ and ${\rm{W}}_Y$ are embedding matrices, $D$ denotes the dimension of bilinear vector, ${\rm{W}}^i_X$/${\rm{W}}^i_Y$ is the $i$-th row in ${\rm{W}}_X$/${\rm{W}}_Y$. Therefore, this expansion clearly shows that each element in bilinear vector after exponential transformation reflects infinity order interactions.

\subsection{X-LAN for Image Captioning}
Recall that our X-Linear attention is a unified attention block, it is feasible to plug X-Linear attention block(s) into image encoder and sentence decoder to capture higher order intra- and inter-modal interactions for image captioning. We next present how to integrate such block(s) into the encoder-decoder structure via our devised X-Linear Attention Networks (X-LAN), as illustrated in Figure \ref{fig:framework}.

\subsubsection{Notation and Training Strategy}
In the standard task of image captioning, we are given an image $I$ to be described with a natural-language sentence $Y_{1:T}$. The sentence $Y_{1:T}=\{{\bf{w}}_1,{\bf{w}}_2,...,{\bf{w}}_T\}$ is a sequence of $T$ words, where ${\rm{w}}_t$ is the textual feature of the $t$-th word. The image $I$ is represented as a set of spatial image region features ${\bf{V}}=\{{\bf{v}}_i\}_{i=1}^N$ by utilizing Faster R-CNN \cite{ren2015faster}. During training, given the ground-truth sentence $Y^\star_{1:T}$ for image $I$, we first train our X-LAN by minimizing the cross entropy loss $L_{CE}(\theta)=-\sum\nolimits_{t = 1}^T {log(p_\theta({\bf{w}}^\star_{t}|Y^\star_{1:t-1}))}$, where $\theta$ denotes the parameters of X-LAN. Next, our X-LAN can be further optimized with sentence-level reward via Self-Critical Sequence Training \cite{rennie2017self}.

\subsubsection{Encoder with X-Linear Attention}
The image encoder is a module that transforms the input set of spatial image region features ${\bf{V}}$ into a series of intermediate states, which are enhanced with the contextual information in between. Here we fully employ our X-Linear attention block(s) to construct the image encoder. As such, the representative capacity of encoded image-level or region-level features are strengthened via capturing higher order intra-modal feature interactions.

Formally, the image encoder in X-LAN is composed of a stack of $(1+M)$ identical layers ($M=3$). Each layer includes two components: X-Linear attention block as in Eq.(\ref{Eq:Eq7-1}) and keys/values updating module as in Eq.(\ref{Eq:Eq7-2}). Specifically, for the first X-Linear attention block, we take the mean-pooled region feature $\hat{\bf{v}}^{(0)}=\overline {\bf{v}} = \frac{1}{N}\sum\nolimits_{i = 1}^N {{\bf{v}}_i}$ as the initial input query, coupled with the initial keys/values (\emph{i.e.}, all region features ${\bf{K}}^{(0)}={\bf{V}}^{(0)}={\bf{V}}$). The output is thus the attended image-level feature $\hat{\bf{v}}^{(1)}$, which will be further fed into the next X-Linear attention block as input query. Meanwhile, the keys/values are updated conditioned on the attended image-level feature $\hat{\bf{v}}^{(1)}$. After that, we repeat the updating process of query and keys/values in $M$ times via the subsequence $M$ stacked layers. Accordingly, by performing feature enhancement via the image encoder with $(1+M)$ X-Linear attention blocks, we can obtain $(1+M)$ output attended image-level features $\{\hat{\bf{v}}^{(m)}\}_{m=1}^{1+M}$. Moreover, we treat the updated values ${\bf{V}}^{(1+M)}$ after the final X-Linear attention block as the enhanced region-level features, which are endowed with the higher order intra-modal feature interactions in between.

\subsubsection{Decoder with X-Linear Attention}
The sentence decoder aims to generate the output sentence conditioned on the enhanced image-level and region-level visual features induced by the image encoder. To further encourage high order inter-modal interactions between visual content and natural sentence, we integrate our X-Linear attention block into attention-based LSTM decoder to perform multi-modal reasoning. In particular, at each decoding time step $t$, we firstly concatenate the mean-pooled region feature $\hat{\bf{v}}^{(0)}$ and all attended image-level features $\{\hat{\bf{v}}^{(m)}\}_{m=1}^{1+M}$, which is further transformed into the global image-level feature $\tilde{\bf{v}}$ through an embedding layer:
\begin{equation}\label{Eq:Eq9}\small
\tilde{\bf{v}}={\rm{W}}_G[\hat{\bf{v}}^{(0)},\hat{\bf{v}}^{(1)},...,\hat{\bf{v}}^{(1+M)}],
\end{equation}
where ${\rm{W}}_G$ is embedding matrix. The input of LSTM is thus set as the concatenation of current input word ${\rm{w}}_t$, the global image-level feature $\tilde{\bf{v}}$, the previous LSTM hidden state ${\bf{h}}_{t-1}$, and the pervious context vector ${\bf{c}}_{t-1}$. After that, we take the output of LSTM ${\bf{h}}_{t}$ as input query of X-Linear attention block, whose keys/values are set as the enhanced region-level features ${\bf{V}}^{(1+M)}$ from image encoder. In this way, the output attended feature $\hat{\bf{v}}_d$ of X-Linear attention block is more representative by capturing the 2$^{nd}$ order interactions between image features and hidden state. Next, we measure current context vector ${\bf{c}}_{t}$ by concatenating the attended feature $\hat{\bf{v}}_d$ with current LSTM hidden state ${\bf{h}}_{t}$, followed with an embedding layer and a Gated Linear Unit (GLU) \cite{dauphin2017language}. Such context vector ${\bf{c}}_{t}$ is finally leveraged for the prediction of next word ${\bf{w}}_{t+1}$ via a softmax layer.

\begin{table*}[t]\scriptsize
    \centering
    \vspace{-0.1in}
    \caption{\small Performance comparisons on COCO Karpathy test split, where B@$N$, M, R, C and S are short for BLEU@$N$, METEOR, ROUGE-L, CIDEr and SPICE scores. All values are reported as percentage (\%). $^{\sum}$ indicates model ensemble/fusion.}
    \vspace{-0.0in}
    \begin{tabular}{l | c c c c c c c c | c c c c c c c c}
        \Xhline{2\arrayrulewidth}
		  & \multicolumn{8}{c|}{\textbf{Cross-Entropy Loss}} & \multicolumn{8}{c}{\textbf{CIDEr Score Optimization}} \\
		                             & B@1  & B@2  & B@3  & B@4  & M    & R    & C     & S    & B@1  & B@2  & B@3  & B@4  & M    & R    & C     & S  \\	
			\hline \hline
LSTM \cite{Vinyals14}            &   -  &   -  &   -  & 29.6 & 25.2 & 52.6 & 94.0  &  -   &  -   &  -   &  -   & 31.9 & 25.5 & 54.3 & 106.3 &  -   \\
SCST \cite{rennie2017self}       &   -  &   -  &   -  & 30.0 & 25.9 & 53.4 & 99.4  &  -   &  -   &  -   &  -   & 34.2 & 26.7 & 55.7 & 114.0 &  -   \\
LSTM-A \cite{yao2017boosting}    & 75.4 &   -  &   -  & 35.2 & 26.9 & 55.8 & 108.8 & 20.0 & 78.6 &  -   &  -   & 35.5 & 27.3 & 56.8 & 118.3 & 20.8 \\
RFNet \cite{jiang2018recurrent}  & 76.4 & 60.4 & 46.6 & 35.8 & 27.4 & 56.5 & 112.5 & 20.5 & 79.1 & 63.1 & 48.4 & 36.5 & 27.7 & 57.3 & 121.9 & 21.2 \\
Up-Down \cite{anderson2017bottom}& 77.2 &   -  &   -  & 36.2 & 27.0 & 56.4 & 113.5 & 20.3 & 79.8 &  -   &  -   & 36.3 & 27.7 & 56.9 & 120.1 & 21.4 \\
GCN-LSTM \cite{yao2018exploring} & 77.3 &   -  &   -  & 36.8 & 27.9 & 57.0 & 116.3 & 20.9 & 80.5 &  -   &  -   & 38.2 & 28.5 & 58.3 & 127.6 & 22.0 \\
LBPF \cite{qin2019look}           & 77.8 &   -  &   -  & 37.4 & 28.1 & 57.5 & 116.4 & 21.2 & 80.5 &  -   &  -   & 38.3 & 28.5 & 58.4 & 127.6 & 22.0 \\
SGAE \cite{Yang:CVPR19}          & 77.6 &   -  &   -  & 36.9 & 27.7 & 57.2 & 116.7 & 20.9 & 80.8 &  -   &  -   & 38.4 & 28.4 & 58.6 & 127.8 & 22.1 \\
AoANet \cite{huang2019attentio}  & 77.4 &   -  &   -  & 37.2 & 28.4 & 57.5 & 119.8 & 21.3 & 80.2 &  -   &  -   & 38.9 & 29.2 & 58.8 & 129.8 & 22.4 \\
      \hline
X-LAN                         & \textbf{78.0} & \textbf{62.3} & \textbf{48.9} & \textbf{38.2} & \textbf{28.8} & \textbf{58.0} & \textbf{122.0} & \textbf{21.9} & 80.8 & 65.6 & 51.4 & 39.5 & \textbf{29.5} & \textbf{59.2} & 132.0 & \textbf{23.4} \\
      \hline
Transformer \cite{sharma2018conceptual}   & 76.1 & 59.9 & 45.2 & 34.0 & 27.6 & 56.2 & 113.3 & 21.0 & 80.2 & 64.8 & 50.5 & 38.6 & 28.8 & 58.5 & 128.3 & 22.6 \\
X-Transformer                         & 77.3 & 61.5 & 47.8 & 37.0 & 28.7 & 57.5 & 120.0 & 21.8 & \textbf{80.9} & \textbf{65.8} & \textbf{51.5} & \textbf{39.7} & \textbf{29.5} & 59.1 & \textbf{132.8} & \textbf{23.4} \\\hline
			& \multicolumn{16}{c}{\textbf{Ensemble/Fusion}} \\ \hline
SCST \cite{rennie2017self}$^{\sum}$ & -   &   -  &   -  & 32.8 & 26.7 & 55.1 & 106.5 &   -  &  -   &  -   &  -   & 35.4 & 27.1 & 56.6 & 117.5 &  -   \\
RFNet \cite{jiang2018recurrent}$^{\sum}$ & 77.4 & 61.6 & 47.9 & 37.0 & 27.9 & 57.3 & 116.3 & 20.8 & 80.4 & 64.7 & 50.0 & 37.9 & 28.3 & 58.3 & 125.7 & 21.7 \\
GCN-LSTM \cite{yao2018exploring}$^{\sum}$& 77.4 & - & - & 37.1 & 28.1 & 57.2 & 117.1 & 21.1 & 80.9 &  -   &  -   & 38.3 & 28.6 & 58.5 & 128.7 & 22.1 \\
SGAE \cite{Yang:CVPR19}$^{\sum}$   &  -   &   -  &   -  &  -   &   -  &   -  &   -   &   -  & 81.0 &  -   &  -   & 39.0 & 28.4 & 58.9 & 129.1 & 22.2 \\
HIP \cite{yao2019hierarchy}$^{\sum}$ & -   &   -  &   - &  {38.0}  &  {28.6}  &  {57.8}  &  {120.3} & {21.4} & -   &   -  &   -  &  {39.1}  &  {28.9}  &  {59.2}  & {130.6}   & {22.3} \\
AoANet \cite{huang2019attentio}$^{\sum}$ & 78.7 & - & - & 38.1 & 28.5 & 58.2 & 122.7 & 21.7 & 81.6 &  -   &  -   & 40.2 & 29.3 & 59.4 & 132.0 & 22.8 \\\hline
X-LAN$^{\sum}$ & \textbf{78.8} & \textbf{63.4} & \textbf{49.9} & \textbf{39.1} & \textbf{29.1} & \textbf{58.5} & \textbf{124.5} & \textbf{22.2} & 81.6 & 66.6 & 52.3 & 40.3 & 29.8 & 59.6 & 133.7 & 23.6 \\\hline
X-Transformer$^{\sum}$ & 77.8 & 62.1 & 48.6 & 37.7 & 29.0 & 58.0 & 122.1 & 21.9 & \textbf{81.7} & \textbf{66.8} & \textbf{52.6} & \textbf{40.7} & \textbf{29.9} & \textbf{59.7} & \textbf{135.3} & \textbf{23.8} \\

		\Xhline{2\arrayrulewidth}
    \end{tabular}
	\vspace{-0.2in}
    \label{tab:COCO}
\end{table*}

\section{Experiments}

\subsection{Dataset and Implementation Details}

All the experiments are conducted on the most popular image captioning benchmark COCO \cite{Lin:ECCV14}. The whole COCO dataset contains 123,287 images, which includes 82,783 training images, 40,504 validation images, and 40,775 testing images. Each image is equipped with five human-annotated sentences. Note that the annotations for official testing set are not provided and the evaluation over that testing set can only be conducted through online testing server. In addition, we adopt the widely adopted Karpathy split \cite{Karpathy:CVPR15} for offline evaluation. There are 113,287 training images, 5,000 validation images, and 5,000 testing images in the Karpathy split. We pre-process all training sentences by converting them into lower case and dropping the words that occur less than 6 times, leading to the final vocabulary with 9,488 unique words.

We leverage the off-the-shelf Faster-RCNN pre-trained on ImageNet \cite{ImageNet} and Visual Genome \cite{krishna2017visual} to extract image region features \cite{anderson2017bottom}. Each original region feature is a 2,048-dimensional vector, which is transformed as the input region feature with the dimension ${D_v}$ = 1,024. Each word is represented as ``one-hot" vector. The dimensions of the bilinear query-key representation and the transformed bilinear feature (${D_B}$ and ${D_c}$) in X-Linear attention block is set as 1,024 and 512, respectively. We stack four X-Linear attention blocks (plus ELU) in the image encoder and the sentence decoder is equipped with one X-Linear attention block (plus ELU). The hidden layer size in LSTM decoder is set as 1,024. The whole image captioning architecture are mainly implemented with PyTorch, optimized with Adam \cite{kingma2014adam}.
For the training stage, we follow the training schedule in \cite{vaswani2017attention} to optimize the whole architecture with cross-entropy loss. The warmup steps are set as 10,000 and the mini-batch size is 40. Since low-rank bilinear pooling may lead to slow convergence rate as indicated in \cite{kim2016hadamard}, we set the maximum iteration as 70 epoches. For the training with self-critical training strategy, as in \cite{rennie2017self}, we first select the initialization model which is trained with cross-entropy loss and achieves best CIDEr score on validation set. After that, the whole architecture is further optimized with CIDEr reward, when the learning rate is set as 0.00001 and the maximum iteration is 35 epoches. At the inference stage, we adopt the beam search strategy and set the beam size as 3.
Five evaluation metrics, BLEU@$N$ \cite{Papineni:ACL02}, METEOR \cite{Banerjee:ACL05}, ROUGE-L \cite{lin2004rouge}, CIDEr \cite{vedantam2015cider}, and SPICE \cite{spice2016}, are simultaneously utilized to evaluate our model.

\begin{table*}[!tb]\scriptsize
  \centering
  \caption{\small Leaderboard of the published state-of-the-art image captioning models on the COCO online testing server, where B@$N$, M, R, and C are short for BLEU@$N$, METEOR, ROUGE-L, and CIDEr scores. All values are reported as percentage (\%).}
  \label{table:leaderboard}
  \vspace{-0.00in}
  \begin{tabular}{l|*{13}{c|}c}
  \Xhline{2\arrayrulewidth}
      \multicolumn{1}{c|}{\multirow{2}{*}{{Model}}} & \multicolumn{2}{c|}{{B@1}} & \multicolumn{2}{c|}{{B@2}} & \multicolumn{2}{c|}{{B@3}} & \multicolumn{2}{c|}{{B@4}} & \multicolumn{2}{c|}{{M}} & \multicolumn{2}{c|}{{R}} & \multicolumn{2}{c}{{C}} \\\cline{2-15}
      \multicolumn{1}{c|}{}&c5 &c40 &c5 &c40 &c5 &c40&c5 &c40&c5 &c40&c5 &c40&c5 &c40 \\\hline
       {LSTM-A} (ResNet-152) \cite{yao2017boosting}     & 78.7 & 93.7 & 62.7 & 86.7 & 47.6 & 76.5 & 35.6 & 65.2 & 27.0 & 35.4 & 56.4 & 70.5 & 116.0 & 118.0  \\\hline
	    {Up-Down} (ResNet-101) \cite{anderson2017bottom} & 80.2 & 95.2 & 64.1 & 88.8 & 49.1 & 79.4 & 36.9 & 68.5 & 27.6 & 36.7 & 57.1 & 72.4 & 117.9 & 120.5  \\\hline
      {RFNet} (ResNet+DenseNet+Inception) \cite{jiang2018recurrent}  & 80.4 & 95.0 & 64.9 & 89.3 & 50.1 & 80.1 & 38.0 & 69.2 & 28.2 & 37.2 & 58.2 & 73.1 & 122.9 & 125.1 \\\hline
			{SGAE} (ResNet-101) \cite{Yang:CVPR19}           & 81.0 & 95.3 & 65.6 & 89.5 & 50.7 & 80.4 & 38.5 & 69.7 & 28.2 & 37.2 & 58.6 & 73.6 & 123.8 & 126.5 \\\hline
			{GCN-LSTM} (ResNet-101) \cite{yao2018exploring}  & 80.8 & 95.2 & 65.5 & 89.3 & 50.8 & 80.3 & 38.7 & 69.7 & 28.5 & 37.6 & 58.5 & 73.4 & 125.3 & 126.5 \\\hline
			{AoANet} (ResNet-101) \cite{huang2019attentio}   & 81.0 & 95.0 & 65.8 & 89.6 & 51.4 & 81.3 & 39.4 & 71.2 & 29.1 & 38.5 & 58.9 & 74.5 & 126.9 & 129.6 \\\hline
			{HIP (SENet-154) \cite{yao2019hierarchy}}                    & 81.6 & \textbf{95.9} & 66.2 & 90.4 & 51.5 & 81.6 & 39.3 & 71.0 & 28.8 & 38.1 & 59.0 & 74.1 & 127.9 & 130.2 \\\hline
			{X-LAN (ResNet-101)}                & 81.1 & 95.3 & 66.0 & 89.8 & 51.5 & 81.5 & 39.5 & 71.4 & 29.4 & 38.9 & 59.2 & 74.7 & 128.0 & 130.3 \\\hline
            {X-LAN (SENet-154)}                 & 81.4 & 95.7 & 66.5 & \textbf{90.5} & 52.0 & 82.4 & 40.0 & \textbf{72.4} & \textbf{29.7} & \textbf{39.3} & \textbf{59.5} & \textbf{75.2} & 130.2 & 132.8 \\\hline
			{X-Transformer (ResNet-101)}                & 81.3 & 95.4 & 66.3 & 90.0 & 51.9 & 81.7 & 39.9 & 71.8 & 29.5 & 39.0 & 59.3 & 74.9 & 129.3 & 131.4 \\\hline
			{X-Transformer (SENet-154)}                 & \textbf{81.9} & 95.7 & \textbf{66.9} & \textbf{90.5} & \textbf{52.4} & \textbf{82.5} & \textbf{40.3} & \textbf{72.4} & 29.6 & 39.2 & \textbf{59.5} & 75.0 & \textbf{131.1} & \textbf{133.5} \\
      \Xhline{2\arrayrulewidth}
  \end{tabular}
  \vspace{-0.15in}
\end{table*}

\subsection{Performance Comparison}

\noindent\textbf{Offline Evaluation.}
Table \ref{tab:COCO} summaries the performance comparisons between the state-of-the-art models and our proposed X-LAN on the offline COCO Karpathy test split. Note that for fair comparison, we report the results for each run optimized with both cross entropy loss and CIDEr Score. Meanwhile, we separately show the performances for single models and ensemble/fused models. In general, our X-LAN consistently exhibits better performances than other single models, which include the non-attention baselines (LSTM, LSTM-A) and attention-based methods (SCST, RFNet, and others). The CIDEr score of our X-LAN can achieve 132.0\% with CIDEr score optimization, which is to-date the best performance without any model ensemble and makes the absolute improvement over the best competitor AoANet by 2.2\%. The performance improvements generally demonstrate the key advantage of exploiting higher and even infinity order interactions via our X-Linear attention block, that facilitate both single-modal feature enhancement and multi-modal reasoning for image captioning. In particular, LSTM-A improves LSTM by emphasising semantic attributes at decoding stage. RFNet and Up-Down further boost the performances by involving attention mechanism that learns to identify selective spatial regions for sentence generation. Moreover, by exploiting rich semantic information in images (e.g., visual relations between objects or scene graph) for sentence generation, GCN-LSTM and SGAE exhibit better performance than Up-Down. Nevertheless, the performances of SGAE are lower than AoANet that enhances conventional visual attention by further measuring the relevance between the attention result and the query. This confirms that improving attention measurement is an effective way to enhance the interaction between visual content and natural sentence and thus boost image captioning. In addition, by integrating X-Linear attention block(s) into encoder and decoder, our X-LAN outperforms AoANet, which demonstrates the merit of mining higher and even infinity intra- and inter-modal interactions. Similar to the observations over single models, an ensemble version of our X-LAN by fusing four models with different initialized parameters obtains better performances than other ensemble models.

To fully verify the generalizability of our X-Linear attention block for image captioning, we include a variant of our X-LAN (named \textbf{X-Transformer}) by plugging X-Linear attention blocks into Transformer based encoder-decoder structure.
Table \ref{tab:COCO} also shows the performance comparison between Transformer and our X-Transformer. Note that here Transformer denotes our implementation of Transformer-based encoder-decoder structure as in \cite{sharma2018conceptual}. Similar to the observations in LSTM-based encoder-decoder structure, X-Transformer boosts up the performances by integrating X-Linear attention blocks into the Transformer-based encoder and decoder.
The performance improvements again demonstrate the advantage of exploiting higher order interactions via our X-Linear attention block for image captioning.

\noindent\textbf{Online Evaluation.}
In addition, we evaluate our X-LAN and X-Transformer on the official testing set by submitting the ensemble versions to online testing server. Table \ref{table:leaderboard} details the performances over official testing images with 5 reference captions (c5) and 40 reference captions (c40). Note that here we adopt two common backbones (ResNet-101 \cite{He:CVPR16} and SENet-154 \cite{hu2018squeeze}) for online evaluation.
The results clearly show that compared to all the other published state-of-the-art systems, our X-LAN and X-Transformer exhibit better performances across most~metrics.

\begin{figure}[!tb]
\vspace{-0.06in}
\centering {\includegraphics[width=0.43\textwidth]{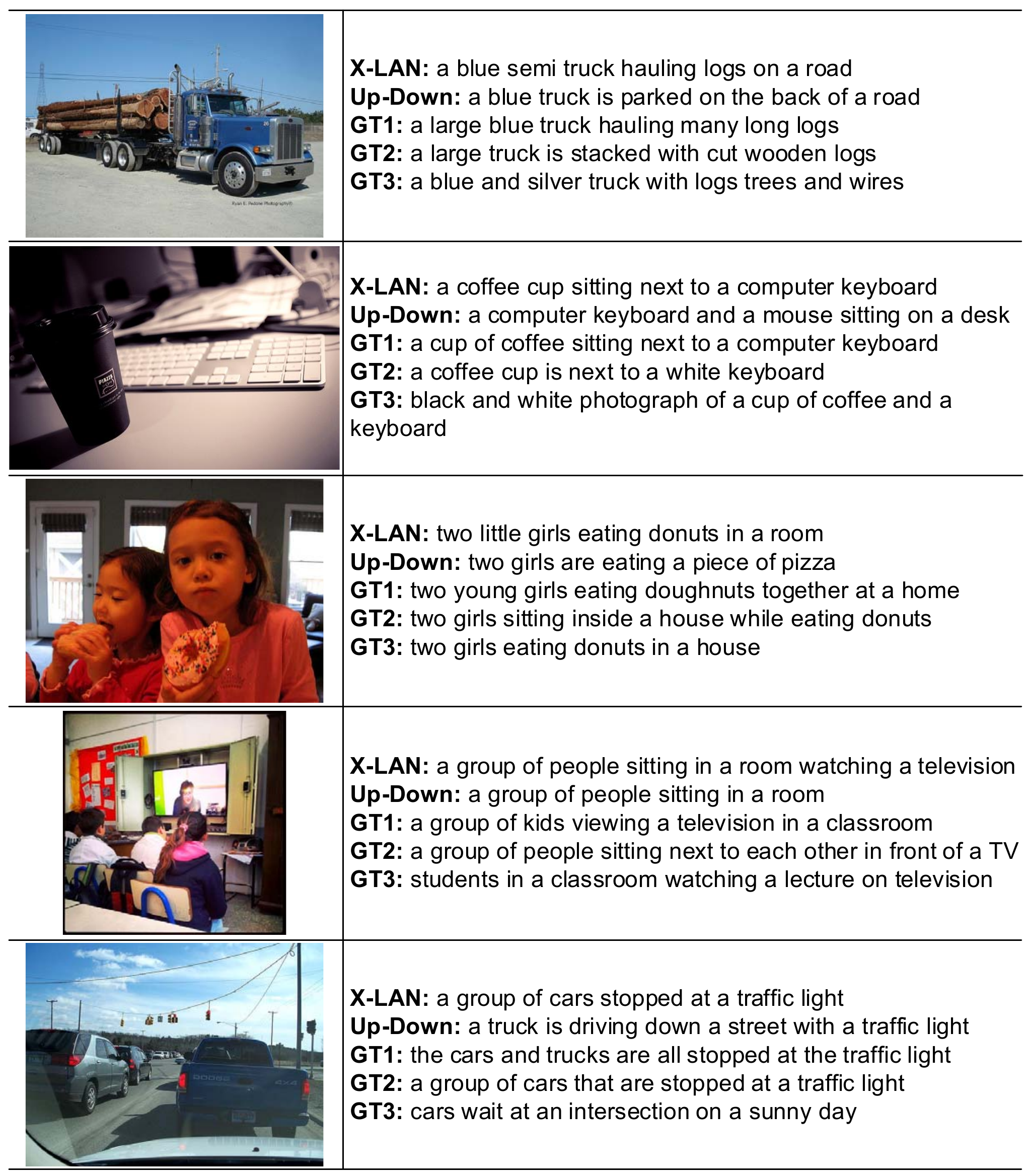}}
\vspace{-0.05in}
\caption{\small Examples of image captioning results by Up-Down and X-LAN, coupled with the corresponding ground truth sentences.}
\label{fig:results}
\vspace{-0.18in}
\end{figure}

\noindent\textbf{Qualitative Analysis.}
Figure \ref{fig:results} showcases several image captioning results of Up-Down and our X-LAN, coupled with human-annotated ground truth sentences (GT). Generally, compared with the captions of Up-Down which are somewhat relevant to image content and logically correct, our X-LAN produces more accurate and descriptive sentences by exploiting higher order intra- and inter-modal interactions.
For example, Up-Down generates the phrase of ``a truck is driving" that is inconsistent with the visual content for the last image, while ``a group of cars stopped" in our X-LAN depicts the visual content more precise.
This again confirms the advantage of capturing the high order interactions among image regions and meanwhile triggering high order interactions between different modalities for multi-modal reasoning via our X-Linear attention block.

\begin{table*}[!tb]\scriptsize
\centering
  \caption{\small Ablation study on the use of X-Linear attention block(s) in image encoder and sentence decoder (optimized with cross-entropy loss), where B@$N$, M, R, and C are short for BLEU@$N$, METEOR, ROUGE-L, and CIDEr. All values are reported as percentage (\%).}
    \label{table:ablation}
\begin{tabular}{l|l|cccccccc}
\Xhline{2\arrayrulewidth}
Image Encoder & Sentence Decoder & B@1 & B@2 & B@3 & B@4 & M & R & C & S \\ \hline\hline
Faster R-CNN & LSTM + Conventional attention & 76.4 & 60.3 & 46.7 & 36.1 & 27.9 & 56.7 & 114.1 & 20.9 \\
Faster R-CNN & LSTM + X-Linear attention & 76.9 & 60.9 & 47.3 & 36.6 & 28.2 & 57.0 & 117.0 & 21.2 \\\hline
Faster R-CNN + 1$\times$X-Linear attention & LSTM + X-Linear attention & 77.3 & 61.5 & 47.9 & 37.1 & 28.5 & 57.3 & 118.2 & 21.6 \\
Faster R-CNN + 2$\times$X-Linear attention & LSTM + X-Linear attention & 77.5 & 61.9 & 48.4 & 37.7 & 28.6 & 57.7 & 119.4 & 21.6 \\
Faster R-CNN + 3$\times$X-Linear attention & LSTM + X-Linear attention & 77.7 & 62.2 & 48.6 & 37.8 & 28.6 & 57.7 & 120.0 & 21.6 \\
Faster R-CNN + 4$\times$X-Linear attention & LSTM + X-Linear attention & 77.8 & 62.3 & 48.7 & 37.8 & 28.6 & 57.8 & 120.4 & 21.6 \\\hline
Faster R-CNN + 4$\times$X-Linear attention (+ELU) & LSTM + X-Linear attention (+ELU) & \textbf{78.0} & \textbf{62.3} & \textbf{48.9} & \textbf{38.2} & \textbf{28.8} & \textbf{58.0} & \textbf{122.0} & \textbf{21.9} \\ \Xhline{2\arrayrulewidth}
\end{tabular}
\end{table*}

\begin{figure*}[!tb]
\vspace{-0.18in}
\centering {\includegraphics[width=0.9\textwidth]{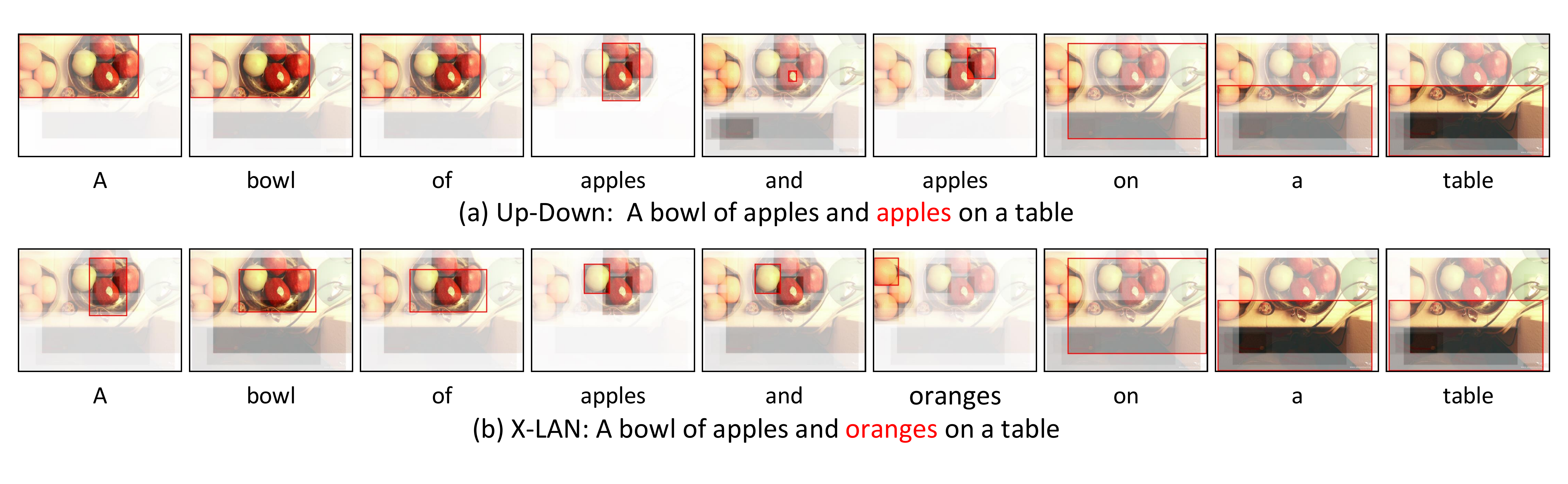}}
\vspace{-0.2in}
\caption{\small The visualization of attended image regions along the caption generation processes for (a) Up-Down and (b) X-LAN. At the decoding step for each word, we outline the image region with the maximum attention weight in red.}
\label{fig:att}
\vspace{-0.22in}
\end{figure*}

\subsection{Experimental Analysis}
\noindent\textbf{Ablation Study.}
To fully examine the impact of X-Linear attention block(s) in both image encoder and sentence decoder for image captioning, we conduct ablation study by comparing different variants of our X-LAN in Table \ref{table:ablation}. We start from a base model which is a degraded version of our X-LAN by simply encoding images with Faster R-CNN and further leveraging LSTM with conventional attention module for sentence generation. This ablated base model obtains similar results to Up-Down. Next, we extend the base model by replacing the conventional attention module with our X-Linear attention block in sentence decoder, which obtains better performances. The results indicate that compared to conventional attention module that only exploits 1$^{st}$ order inter-modal interaction, our X-Linear attention block enhances the capacity of multi-modal reasoning via the modeling of higher order interactions. Furthermore, in order to show the relations between performance and the number of stacked X-Linear attention blocks in image encoder, we include a series of variants by integrating one more X-Linear attention blocks into encoder. In general, stacking more X-Linear attention blocks in image encoder can lead to performance improvements. That basically validates the effectiveness of modeling high order intra-modal interactions between image regions in encoder. However, no further significant performance improvement is observed when stacking four blocks. We speculate that the increased parameters by stacking more blocks might result in overfitting, which somewhat hinder the exploitation of more higher order interaction in this way. Recall that instead of stacking blocks to capture more higher order interactions, we present a simple but effective solution to enable even infinity order feature interactions by equipping X-Linear attention block with ELU in a parameter-free fashion. As such, when upgrading our X-Linear attention block with ELU in both encoder and decoder, a larger performance gain is attained.

\noindent\textbf{Attention Visualization.}
In order to better qualitatively evaluate the generated results with X-Linear attention block, we visualize the evolutions of attended image regions along the caption generation processes for Up-Down and X-LAN in Figure \ref{fig:att}. We can observe that Up-Down sometimes focus on the irrelevant image region whose corresponding object should not be generated at that time step. For instance, at the 6$^{th}$ decoding step for Up-Down, the region containing irrelevant object ``apples" is attended, which misleads decoder to produce ``apples" again. Instead, by exploiting higher order feature interactions for multi-modal reasoning via X-Linear attention block, our X-LAN consistently focuses on the correct regions for image captioning.

\section{Conclusions}
We present a novel unified X-Linear attention block for image captioning, which models the 2$^{nd}$ order interactions with both spatial and channel-wise bilinear attention. The higher and even infinity order feature interactions can be readily modeled via staking multiple X-Linear attention blocks and equipping the block with Exponential Linear Unit (ELU). To verify our claim, we devise X-Linear Attention Networks (X-LAN) by integrating X-Linear attention block(s) into image encoder and sentence decoder to exploit higher order intra and inter-modal interactions for image captioning.
Extensive experiments conducted on COCO dataset demonstrate the efficacy of X-Linear attention block and X-LAN.
More remarkably, we obtain new state-of-the-art performances on this captioning dataset with X-LAN.

{\small
\bibliographystyle{ieee_fullname}
\bibliography{egbib}
}

\end{document}